\documentclass{article}

\usepackage{arxiv}

\usepackage[utf8]{inputenc} % allow utf-8 input
\usepackage[T1]{fontenc}    % use 8-bit T1 fonts
\usepackage{hyperref}       % hyperlinks
\usepackage{url}            % simple URL typesetting
\usepackage{booktabs}       % professional-quality tables
\usepackage{amsfonts}       % blackboard math symbols
\usepackage{nicefrac}       % compact symbols for 1/2, etc.
\usepackage{microtype}      % microtypography
\usepackage{lipsum}
\usepackage{graphicx}
\usepackage[table]{xcolor}
\usepackage{tabularx}
\usepackage{enumerate}
\usepackage{amsmath}
\usepackage{scrextend}
\usepackage{float}
\usepackage{booktabs, caption, makecell}

\usepackage{threeparttable}
\usepackage{algorithm}
\usepackage{algpseudocode}
\usepackage{url}
\usepackage{hyperref}
\usepackage{afterpage}
\usepackage{tikz}
\usepackage[T1]{fontenc}
\usepackage{ebgaramond}
\usepackage{amsfonts}
\usepackage{multirow}

\def\cca#1{\cellcolor{black!#1}\ifdim #1pt > 50pt\color{white}\fi{#1}}

\title{Deep Multi-Shot Network for modelling Appearance Similarity in Multi-Person Tracking applications}

\author{
 M. J. Gómez-Silva\\
 Intelligent Systems Lab (LSI) Research Group, \\
 Universidad Carlos III de Madrid, \\ 
 Leganés, Madrid, Spain \\
 \texttt{magomezs@ing.uc3m.es}
}

\begin{document}
\maketitle
\begin{abstract}
The automatization of Multi-Object Tracking becomes a demanding task in real unconstrained scenarios, where the algorithms have to deal with crowds, crossing people, occlusions, disappearances and the presence of visually similar individuals. 
In those circumstances, the data association between the incoming detections and their corresponding identities could miss some tracks or produce identity switches.
In order to reduce these tracking errors, and even their propagation in further frames, this article presents a Deep Multi-Shot neural model for measuring the Degree of Appearance Similarity (MS-DoAS) between person observations. This model provides temporal consistency to the individuals' appearance representation, and provides an affinity metric to perform frame-by-frame data association, allowing online tracking.
The model has been deliberately trained to be able to manage the presence of previous identity switches and missed observations in the handled tracks.
With that purpose, a novel data generation tool has been designed to create training tracklets that simulate such situations.
The model has demonstrated a high capacity to discern when a new observation corresponds to a certain track, achieving a classification accuracy of 97\% in a hard test that simulates tracks with previous mistakes. 
Moreover, the tracking efficiency of the model in a Surveillance application has been demonstrated by integrating that into the frame-by-frame association of a Tracking-by-Detection algorithm.
\end{abstract}

% keywords can be removed

\section{Introduction}
\label{intro}
%MOT
Multi-Object Tracking (MOT) task consists of visually finding the location of multiple individuals from their visual measurements and conserving their identities in an image sequence.

%automatization for surveillance 
In the context of Intelligent Surveillance Systems, the automatization of the MOT task is essential to manage the huge amount of data captured from a large-scale distributed network of cooperative sensors and consequently, to automatically monitor multiple individuals in wide areas.
%association
This automatization relies on the proper association between the consecutive observations of each individual along a surveillance sequence.  

%movement problems
In real unconstrained and crowded scenarios, the tracking of multiple individuals is hampered by a wide variety of challenging situations: fast-moving people or moving camera platforms, presence of crowds, crossing people, people with changing-trajectories, partially or total occlusions along short or long term, people disappearing from the monitored area, or new individuals entering in the field of view of the surveillance camera.

The performance of the data association process substantially depends on the design of the person representation and on the formulation of the cost function. This function is a metric to measure the cost of assigning a certain identity to a certain detection. % from the current frame.
Consequently, data association methods based on only motion cues or targets' dynamics, e.g. \cite{mclaughlin2015enhancing}, are not able to handle agents with varying trajectories. This circumstance boosts the research on modelling individuals' appearance to improve the performance of online methods.
 
%problem of unknown people
In addition, no information about the agents appearing in the scene is known in advance. Given the unpredictable nature of the surveillance task, an essential capacity for MOT algorithms is the versatility to be applied to any unknown individual, who must be recognised among a high number of observations.

%doas
To achieve that, instead of learning a number of specified patterns for each one of the tracked agents, e.g.\cite{yang2016temporal}, this article proposes the design of a unique deep neural model. The proposed network jointly models the appearance features of multiple person detections and an affinity metric to compare them, which results in the measurement of the Degree of Appearance Similarity (DoAS) between the person images.
This model identifies the affinity between different images of the same person, allowing the tracking of multiple people using the same model for all of them. 
In that way, unlike online-learning models approaches, the developed method does not require previous knowledge about the scene and neither a large number of frames to learn a robust model for an agent's appearance. %The proposed agents' identification presents the same accuracy performance from the first frame of the tracking.
%consistency
%Although deep learning has been proven to provide successful results in many fields of application, its capability to learn a unique appearance model able to represent any anonymous individual in a scene, has not been sufficiently exploited yet. 

The recognition of a person by means of an appearance neural model presents an intrinsically unbalanced nature, given the lack of data about the people to identify and the huge number of possible false assignments with surrounding agents. This results in the collapse and over-fitting of the neural model.
For that reason, a novel formulation to generate the proper training data to feed the model is proposed in this work. %Besides, after training the DoAS model,  this has been integrated into a multi-modal cost function where also motion predictions are considered.

Once the model is trained and integrated into a data association process, its performance in complex scenes could produce some identities switches, i.e. the association of incorrect identities to some detections. 
After an identity switch, a different person’s track is associated with an agent in further iterations, making very difficult the correction of such error. 

With the aim of avoiding identity switches or dealing with the consequences of a previous mismatching, and in order to avoid the error propagation in further frames, temporal consistency has been implicitly added to the model through a novel contrastive network architecture design.
This follows a Multi-Shot recognition approach, whose core is a Long Short Term Memory (LSTM) cell. 

A Deep Convolutional Neural Network has been modelled to render the appearance of the individuals through a feature array.
The obtained features for a certain individual at different frames are related by the LSTM cell, providing the global appearance feature for a certain track. This is compared with the new observations by the proposed model. The result is a contrastive metric, hereinafter called Multi-Shot DoAS. In that way, every detection is compared by a model that considers not only the last saved observations but also those from previous frames. 

%ms_doas proposal
%contribution
Therefore, this article presents a novel neural model to measure the Multi-Shot Degree of Appearance Similarity (MS-DoAS) between person images to perform the association of individuals’ observations through different frames in a Multi-Object Tracking algorithm. 
The main contributions of the proposed method are:

\begin{enumerate}[i.]
	\item Design of a novel Deep Neural Network architecture for performing Multi-Shot recognition of any unknown individual. This model relies on the temporal consistency of the agent’s appearance by analysing the visual features measured in previous frames. 
	\item Formulation of a training process that makes the model able to face complex real surveillance situations, including short-term disappearances of the agents, missed detections, and occlusions. Moreover, the resultant model is also able to deal with previously failed associations, preventing from further propagation of the identities mismatching. 
	These capabilities have been acquired by training it on a variate set of tracklets (fragments of tracks), especially generated with such purpose by deliberately introducing temporal steps between some captures, as well as, intruder detections\footnote {The tracklets generation tool has been implemented as a set of C++ functions, which are publicly available under http://github.com/magomezs/dataset\_factory/tree/master/data\_factory\_from\_mot}. 
	\item Integration of the proposed model in the data association process of an online Multi-Object Tracking algorithm. The affinity measure given by the proposed model, MS-DoAS, has been used, together with other motion cues, as part of a multi-modal cost function. 
\end{enumerate}

%design and learning doas model for any individual
%robust consistency through dedicated training set
%integration on an online tracking

%Therefore, the MS-DoAS cue is provided by a unique pre-trained model, without requiring previous knowledge about the scene. 
The proposed model provides temporary consistency by modelling the agents’ appearance with an LSTM network which is fed by features from previous frames.  In that way, the propagation of punctual association mistakes is avoided, without requiring extending the association process through multiple future frames, as batch methods do. Hence, the proposed model allows an online tracking algorithm, with a frame-to-frame assignment.  

The effectiveness of the proposed model has been proved by measuring its recognition capacity over multiple and variate test sets, and also by evaluating the final performance of an MOT algorithm where the model is integrated.    

%structura
The rest of the article is structured as follows: the second Section presents a review of the existing related works. Section 3 describes the proposed Re-Id neural model, and Section 4 the developed learning algorithm to train it. Finally, Section 5 and 6 present the obtained experimental results and some concluding remarks, respectively.

\section{Related Work}
\label{related}

%Nowadays, the development of Single-Object Tracking algorithms is considerably mature, unlike  methods, which are still . 
Although Multi-Object Tracking (MOT) methods have been reviewed intensively, \cite{bernardin2008evaluating}, it remains a challenging problem under development.
MOT has become a branch of research deeply studied by the scientific community due to its prominent application to Intelligent Surveillance Systems, ISS, since many other applications, such as behaviour analysis, rely on the tracking performance. 

%Indeed, in the last decades, the development of  ISS has been boosted by the necessity of automating and optimising the management of the visual data captured by a continuously rising number of cameras, installed for security purposes. 
%In that respect, tracking multiple people is one of the essential modules of an  {ISS}, since 

Furthermore, Multi-Object Tracking in video sequences is also widely used in other military and civil applications, such as sports players tracking and analysis \cite{liu2013tracking}, biology \cite{meijering2009tracking}, robot navigation \cite{ess2008mobile}, and autonomous driving vehicles \cite{ess2009improved}. 

%tracking by detection, parts
In the literature, tracking problem is commonly solved by selecting a detector and feeding a tracker with it, resulting in a wide range of approaches, which have long been encompassed under the paradigm called ``tracking-by-detection''. 

%association, parts
Once a set of reliable detections is collected, the task of the tracker translates into a data association problem for determining the correspondence of detections across frames. Therefore, the data association consists of finding the correct assignment between the detections at every new frame and their corresponding identities. Identity is given to every trajectory that describes the path of an individual instance over time, hereinafter called agent.

%In the past decades, many research efforts have been focused on improving the data association techniques \cite{adam2006robust,kuo2011does,shu2012part,zhang2008global}. 
Data association methods are mainly composed of a cost function, which measures the cost of assigning a certain identity to a detected person, and an optimization algorithm, which is in charge of seeking the assignment that minimizes the cost function. Therefore, independently from the association mechanism, a significant part of the final Multi-person tracking performance relies on the proper formulation of the cost metric, whose is limited, in turn, by the person representation design.

%modeling people
Some of the most commonly used features are related to individuals’ motion, such as location, or velocity, and even the interactions between agents. 
Trajectories have been typically treated as state-space models, like in Kalman \cite{kalman1960new} or particle filters \cite{gordon1993novel}. 
Moreover, in \cite{bae2014robusttracklet,bae2014robust}, trajectories are clustered as a mean to learn motion patterns. 
Furthermore, another approach is to develop more complex motion models to better predict future trajectories. For instance, Fan et al. \cite{fan2010human} used Deep Convolutional Neural Networks (DCNN) to predict the location and scale of an individual for tracking.

%Knowing the likely position of targets in future frames will reduce the search space and hence increase the appearance model accuracy. 
%In addition, recently, the individuals\textquotesingle visual appearance is also widely exploited in tracking systems.
%These relevant cues about motion or appearance are usually encompassed in a model to provide a powerful modelled representation of the tracked people and their new detections

However, in crowded scenes, a location or motion-based online association method could find problems to deal with changing-trajectory and crossing agents. 
There is a vast number of works that exploit appearance information to solve data association and to overcome the dependency on the motion cues. In those cases, a primary task in people tracking is converting raw pixels into higher-level representations.

Some simple appearance models are based on extracting appearance information from the object pixels using hand-crafted features, including colour histogram \cite{le2016long,tang2016multi} and texture descriptors \cite{chen2015multitarget,zhang2015tracking}.
%, which are the most popular representation for appearance modelling in  {MOT}. 

Other approaches use covariance matrix representation, pixel comparison representation and SIFT-like features, or pose features \cite{nam2016learning}. For instance, in \cite{shu2012part} an edgelet-based part model for describing the appearance of objects is presented.

Recently, Deep Convolutional Neural Networks have been used for modelling appearance by learning high-level features, e.g. \cite{held2016learning,leal2016learning,zhai2018deep}. For example, in \cite{kim2015multiple} the feature extraction is directly learnt by using a convolutional pipeline that can be completely trained on a vast number of samples.

Other tracking algorithms get improvements by means of modelling every tracked agent independently, e.g.\cite{bae2014robust,yang2016temporal}. 
Since there is no previous knowledge about the people to track, the dedicated models are trained online. 
%The work presented in \cite{breitenstein2011online} train a classifier for each target and use the classification score for Greedy data association or Particle filtering.  Song et al. \cite{song2008vision} train appearance models online for individual trajectories when they are isolated, and then they are used to disambiguate from other trajectories in difficult situations like occlusions or interactions.
The drawback of these approaches is that a certain time is needed until the online learning catches enough number of samples of a person to learn a reliable pattern. 

%affinity
On the other hand, many works explicitly learn affinity or similarity metrics from data, in order to compare two observations, e.g. \cite{leal2016learning}. 
%Indeed, the data association process relies on affinity scores, which are based on measuring the similarity between the features of both sets, agents and detections, to obtain their matching cost. 
%For instance, in \cite{leal2016learning} a Siamese neural network is trained to compare the appearance of two detections and combine this with spatial and temporal differences in a boosting framework. 
These works are characterised by the use of a cost metric in their tracking formulation once the metric has been learnt, but they do not consider the actual inference model during the learning phase.

%flow methods
The recent trend in Multi-Target tracking is the integration of the people features learning into the association scheme method. This approach is applicable to batch association methods, a.k.a. offline methods, such as multi-dimensional alignment algorithms \cite{reid1979algorithm} and network flow-based methods \cite{schulter2017deep}. 
%batch
Batch association methods provide temporal context through sets of future observations, allowing for robust predictions. 

For example, Multi Hypothesis Tracking method can be extended to include online learned discriminative appearance models for each track hypothesis \cite{kim2015multiple}. 
%Online discriminative appearance modelling is a standard method for addressing appearance variation \cite{smeulders2014visual}. 
On the other hand, in \cite{schulter2017deep}, features for Network Flow-based data association are learnt via back-propagation, by expressing the optimum of a smoothed network Flow problem as a differentiable function of the pairwise association costs. 

Furthermore, many of the research efforts focused on reducing the tracking errors, exploit the temporal consistency by the extraction of people tracklets, i.e. short object tracks. Unfortunately, the availability of reliable tracklets cannot be guaranteed due to the propagation of mistakes.
This effect is pronounced in network flow-based association methods due to their limited capacity to model complex appearance changes. 
An alternative is to define pairwise costs between tracklets that can be reliably computed, \cite{shitrit2014multi}. 
%For instance, Li et al. \cite{li2009learning} present a hierarchical association approach where increasingly longer tracklets are combined into trajectories. The similarity between tracklets is learned by a boosting formulation from several hand-crafted inputs including the length of trajectories and colour histograms. 

In tracklet association, discriminative appearance models are trained with the aim of learning an improved affinity score function, e.g. \cite{bae2014robusttracklet,yang2012online}. 
%several works  train  In Brendel et al. \cite{brendel2011multiobject}, tracklets association is solved by the MWIS method. Kim et al. \cite{kim2015multiple} also adopt  {MWIS} but follow the classical formulation presented in \cite{papageorgiou2009maximum} and focus on the incorporation of appearance modelling.
However, these are batch methods, which perform multi-frame generalization using tracklets or even the whole sequence at once and on a hierarchical global data association \cite{zhang2008global}, where all the detections are gradually connected after these have been collected for a huge set of frames. Therefore, batch methods rely on future observations and for that reason, they are not applicable in real-time vision systems, where a frame-by-frame association, called online association, is needed.

%%lstm
On the contrary, other methods add temporary consistency to the data association process by using Long Short-Term Memory (LSTM) models. Subsequently, the pairwise terms, which relate two observations, can be weighted by offline trained appearance templates \cite{shitrit2011tracking} or a simple distance metric between appearance features \cite{zhang2008global}. 
For instance, in \cite{sadeghian2017tracking} an LSTM model which learns to predict similar motion and appearance patterns is presented.

%offline

Modelling appearance with LSTM cells in an offline learning process and using the obtained models into an online data association method brings together the advantages of allowing a real-time algorithm with temporal consistency, as the work presented in this article demonstrates. This approach does not require any knowledge about individuals and neither time to adapt the model to them, and any unknown agent can be tracked.

\section{Degree of Appearance Similarity Model}

%presentación del modelo
With the aim of exploiting the visual appearance of a target individual to track him/her among multiple people, an appearance affinity model has been developed. 

The differences in visual appearance between a certain agent (tracked identity) and a detection is taken as an affinity cue in their matching cost formulation.
Instead of modelling a specific individual’s appearance pattern, a universal model has been designed to predicts whether the images correspond to the same person or not.

Therefore, a comparative metric has been trained to measure the Degree of Appearance Similarity (DoAS) between captures. This has been formulated as a pair-wise binary classification problem to discriminate between groups of images belonging to the same person, or corresponding to different people, which are called positive and negative tracklets, respectively.

\subsection{Multi-Shot DoaS Architecture}

\begin{figure}[H]
	\centering
\includegraphics[width=\textwidth]{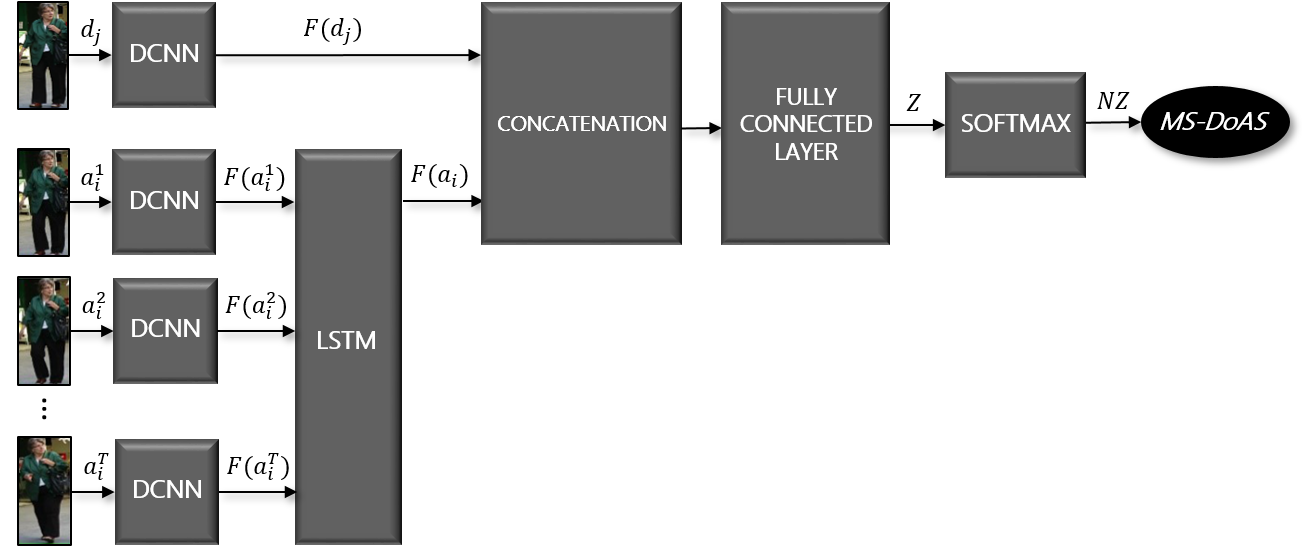}
	\caption{Computation of Multi-Shot Degree of Appearance Similarity, MS-DoAS, between a detection, $d_i$, and an agent, $a_i$.}
	\label{fig:ms_doas_computation}
\end{figure}

%intro to the use of MUlti-shot doas
The Multi-Shot DoAS model, MS-DoAS, measure the appearance affinity between a certain detection, $d_i$, and a certain agent $a_i$, and its computation is rendered by the scheme in Fig. \ref{fig:ms_doas_computation}.

%scheme

%variables description
This model follows a multi-shot recognition scheme since a certain detection, captured in the current frame, $f$, is compared with the visual appearance of a certain agent, not only in the previous iteration, $f-1$, but in $T$ previously captured images. This Multi-Shot recognition approach provides temporary consistency to obtain an accurate prediction. 

Due to certain occlusions, or temporary disappearances, a certain agent could not be detected in consecutive frames. The available captures feed the model in inverse order of acquisition. So, the number, $f_1$, of the frame where the representation $a_i^t$ of the agent was acquired is always higher than the number, $f_2$, of the frame where the next representation $a_i^{(t+1)}$ was captured, $f_1>f_2$. Moreover, all these frames are previous to that where the compared detection was found. 
 
%scheme description
The MS-DoAS metric has been computed by modelling the appearance of each compared individual in a feature array. The appearance of the query detection is rendered by a feature array, $F(d_j)$, computed by a pre-trained DCNN.

Analogously, the appearance of each one of the $T$ previously acquired representations of the query agent, $a_i^t$, is rendered by a feature array, $F(a_i^t)$, previously computed by the same pre-trained DCNN in the frame where it was captured. 
The saved features about the agent are used as the inputs of a pre-trained Long-Short-Term-Memory (LSTM) cell,  which provides a general feature for the agent,  $F(a_i)$.

Subsequently, a second group of neural layers are used to model the affinity cue that contrasts the individuals' appearance features.
Firstly, $F(d_j)$ and $F(a_i)$ are concatenated to feed a fully connected (FC) layer, which has been also pre-trained. This FC layer is used as a binary classification function, which performs the optimal weighting of the elements of the features $F(d_j)$ and $F(a_i)$ and returns a pair of outputs ($K=2$) whose values, $(Z_{0}, Z_{1})$, are representative of the dissimilarity and similarity classes. Finally, a Softmax function, $\sigma$, defined by Eq. \ref{eq:nz_n}, normalises these values in the range $[0, 1]$.

\begin{equation}
\label{eq:nz_n}
NZ_n=\sigma (Z_n) = \frac{e^{Z_n}}{\sum_{k=0}^{K-1}e^{Z_k}}, \forall n \in [0,1]\in \mathbb{Z}  
\end{equation}

Due to the contrastive essence of this pair-wise approach, 
%the proposed affinity model measures the Degree of Appearance Similarity between a tracked agent and a new detection. So 
the first normalised output, $NZ_{0}$, returns the probability of that the agent, $a_i$, and the measurement, $d_j$, do not form a correct match, and the second, $NZ_{1}$, the probability that they form a correct match. Therefore, $NZ_{1}$ is taken as the MS-DoAS between the agent, $a_i$, and the detection, $d_j$.

\subsection{Features computation}
\label{features}

Instead of directly compare the raw images, the comparison is performed from their representative feature arrays. Therefore, it is necessary to model an embedding, $F(I)$, to map an input image, $I$, to a feature space, such that the distance between samples rendering the same person is smaller than that between different people in that feature space.

In order to deal with partial-term occlusions, after which the representation of a person changes, deep learning has been used to automatically find the most salient features of the individuals’ appearance. Hence, the feature embedding has been modelled by a Deep Convolutional Neural Network (DCNN). Therefore, the feature representation for an image, $F_W(I)$, is given by the output of the DCNN, which depends on its weights values, $W$.

%architecture
Concretely, the used DCNN model follows an adapted version of the VGG$11$ architecture, presented as the A version of a set of Very Deep CNNs in \cite{simonyan2014very}. The layers specifications of the proposed VGG$11$-based embedding are listed in Table \ref{tab:vgg11}.

%\renewcommand\arraystretch{0.7} \setlength\minrowclearance{4pt}
\begin{table}[H]
	\centering
	\caption{Structure of the used VGG$11$-based model. The input and output sizes are described in $\#rows$ x $\#cols$ x $\#filters$; the kernel, in $\#rows$ x $\#cols$ x $\#filters$, $stride$, or $\#outputs$ for FC layers.}
	\label{tab:vgg11}       % Give a unique label
	\begin{tabular}{|l|l|l|l|}
		\hline%\noalign{\smallskip}
		Layer &	Input size	& Output size	& Kernel  \\ \hline \hline
		Conv-1-1 &	$128$ x $64$ x $3$ &	$128$ x $64$ x $64$ &	$3$ x $3$ x $3$ \\  \hline
		Pool-1 &	$128$ x $64$ x $64$ &	$64$ x $32$ x $64$  &	$2$ x $2$ x $64$, $2$\\  \hline
		Conv-2-1 &	$64$ x $32$ x $64$ &	$64$ x $32$ x $128$ &	$3$ x $3$ x $64$\\   \hline
		Pool-2 &	$64$ x $32$ x $128$ &	$32$ x $16$ x $128$ &	$2$ x $2$ x $128$, $2$\\  \hline
		Conv-3-1 &	$32$ x $16$ x $128$ &	$32$ x $16$ x $256$ &	$3$ x $3$ x $128$\\  \hline
		Conv-3-2 &	$32$ x $16$ x $256$ &	$32$ x $16$ x $256$ &	$3$ x $3$ x $256$\\  \hline
		Pool-3 &	$32$ x $16$ x $256$ &	$16$ x $8$ x $256$ &	$2$ x $2$ x $256$, $2$\\  \hline
		Conv-4-1 &	$16$ x $8$ x $256$ &	$16$ x $8$ x $512$ &	$3$ x $3$ x $256$\\  \hline
		Conv-4-2 &	$16$ x $8$ x $512$ &	$16$ x $8$ x $512$ &	$3$ x $3$ x $512$\\  \hline
		Pool-4 &	$16$ x $8$ x $512$ &	$8$ x $4$ x $512$ &	$2$ x $2$ x $512$, $2$\\  \hline
		Conv-5-1 &	$8$ x $4$ x $512$ &	$8$ x $4$ x $512$ &	$3$ x $3$ x $512$\\  \hline
		Conv-5-2 &	$8$ x $4$ x $512$ &	$8$ x $4$ x $512$ &	$3$ x $3$ x $512$\\  \hline
		Pool-5 &	$8$ x $4$ x $512$ &	$4$ x $2$ x $512$ &	$2$ x $2$ x $512$ ,$2$\\  \hline
		FC-6 &	$4$ x $2$ x $512$ &	$1$ x $1$ x $4096$ &	$4096$\\  \hline
		FC-7 &	$1$ x $1$ x $4096$ &	$1$ x $1$ x $4096$ &	$4096$\\  \hline
		FC-8 &	$1$ x $1$ x $4096$ &	$1$ x $1$ x $1000$ &	$1000$\\  \hline
	\end{tabular}
\end{table}

VGG$11$ presents eight convolutional layers, three fully connected layers and a SoftMax final layer. 
The SoftMax layer has been removed to get a feature array as output instead of a classification probability value. Hence, its output is a point in the feature space represented by a $1000$-dimensional array $F(I) \in \mathbb{R}^n$ ($n=1000$). Moreover, the input size used in \cite{simonyan2014very} has been modified to adapt its value to the person detections proportions. Therefore, the input of the proposed DCNN is an RGB image of a fixed size, which has been set 64x128 pixels.
All hidden layers are provided of a Rectified Liner Unit, ReLU, \cite{krizhevsky2012imagenet}, as activation function.

%pretained weights (references transfering, overfiting)
This neural network has been trained following the Siamese model, which can perform the discrimination of the pairs of samples in two well-differentiated groups, positive and negative pairs. This discrimination has been accentuated by the use of the Normalised Double-Margin-based Contrastive Loss Function\footnote{The Normalised Double-Margin-based Contrastive Loss function has been implemented in a Caffe python layer, which is publicly available under http://github.com/magomezs/N2M-Contrastive-Loss-Layer}, formulated in \cite{gomez2017deep}.

The Pair-based Mini-Batch Gradient Descent Learning Algorithm, presented in\cite{gomez2019balancing}, has been conducted to learn the network weights. The training data has been generated from the MOT17 dataset of surveillance sequences. 
A data generation tool was used to extract people detections from the sequences, and subsequently, the detections are combined to create a huge number of training pairs, by means of using the balancing data method\footnote{The data generation tool, which includes a data balancing method, is composed of a set of C++ functions that are publicly available under http://github.com/magomezs/dataset\_factory/tree/master/data\_factory\_from\_mot}, presented in \cite{gomez2019balancing}.

Once the neural network was trained, this was used to compute the appearance feature of a person’s image\footnote{The C++ class needed to interpret the network architecture and its pre-trained weights is publicly available under $https://github.com/magomezs/feature\_computation$}.

%---------------------------------------------------------------------------------

\section{Learning Algorithm}
\label{learning}

\begin{figure}[H]
	\centering
	\includegraphics[width=0.65\textwidth]{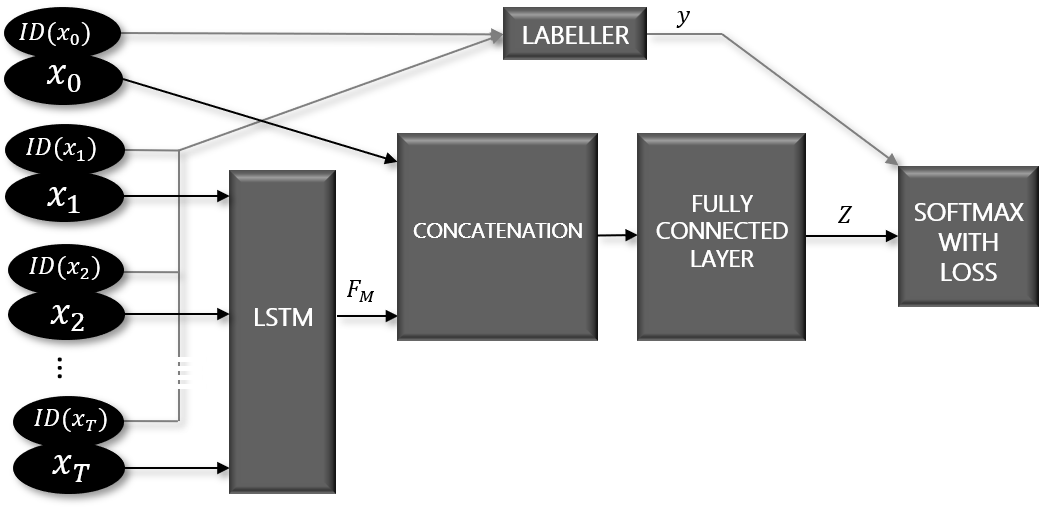}
	\caption{ {MS-DoAS} learning architecture.}
	\label{fig:lstm_training}
\end{figure}

The training architecture used to learn the parameters of the LSTM cell and the FC layer that allow measuring the MS-DoAS is rendered in Fig. \ref{fig:lstm_training}. 

Each training sample, ~$X_i=(x_0,\ldots x_T)$ is formed by an array (with zero-based indexing) of features, called feature tracklet, whose size is $T+1$, where $T$ is the size of the memory of the LSTM cell.
Each tracklet element, $x_n$, corresponds to a feature computed from an image that was taken from a frame whose number is given by $f(x_n)$.

By computing the features in an offline process, previous to the learning, the training time is highly reduced. 
The number of possible training tracklets created from a given set of images is much larger than the size of that set of images. For that reason, computing the features of all the images of the set before forming tracklets combinations is much more efficient, as long as the computational time is concerned.

Because the neural model is learnt by a supervised training process, every feature of a tracklet is accompanied by the identification number of the person that it is rendering, $ID(x_n)$.  A tracklet is considered as positive $(y=1)$ if its first feature, $x_0$, corresponds to the same individual than the rest of features of the tracklet, and it is considered as negative $(y=0)$ in the opposite case, as Eq. \ref{eq:tracklet_labeller} defines. The identity rendered by the $T$ last features of the tracklet is given by the identity represented by the most of its components (the mode, $Mo$) since some intruders (with different $id$) can be added to the tracklet to simulate failed associations.
\begin{equation}
\label{eq:tracklet_labeller}
y=\begin{cases}
1 & if ID(x_0)= Mo(ID(x_n)) \\
0 & if ID(x_0)\neq  Mo(ID(x_n))
\end{cases}
\end{equation}

A feature tracklet-based version of the Mini-Batch Gradient Descent algorithm has been implemented to train the presented neural model, and its main procedures, along with the learning iterations, $it$, are described by Algorithm \ref{al:trackletsSGD}. This algorithm is based on the use of the cross-entropy loss, $f_{CE}$ to compute the loss function, $f_L$, as Eq.\ref{eq:f_l} and \ref{eq:f_CE} define, on its forward propagation and its derivatives, Eq.  \ref{eq:loss_softmax_derivative}, on the backward propagation. 
These partial derivatives are finally used to update the weights using the Adagrad optimisation method, \cite{duchi2011adaptive}.

\begin{algorithm}[H]
	%\label{al:tracklets}
	\caption{Feature Tracklet-based Mini-Batch Gradient Descent Learning Algorithm.}\label{al:trackletsSGD}
	\begin{algorithmic}[1]
		\Require Batch of feature tracklets, $X^{it}=\left\{ X_i^{it} \right\}$.
		\Ensure The network parameters $W^{IT}={W_j^{IT}}$
		\State $W^0=\left\{w_j^0\right\}$
		\While{it<IT}
		\State $it \leftarrow it+1$
		\State  $\frac{\partial f_{L}}{\partial W_j} =0 $;
		\For {\textbf{all} training tracklet $X_i$ of the batch set $X^{it}$}
		\State Calculate $Z_i=(Z_{i,0}, Z_{i,1})$ by forward propagation; 
		\State Calculate $\sigma(Z_{i,n})$ by Eq. \ref{eq:nz_n};
		\State Calculate $ f_{CE}(W^{it}; X^{it}_i)$ by Eq.\ref{eq:f_CE};
		\EndFor
		\State Calculate $f_{L} (W^{it}; X^{it})$  by Eq. \ref{eq:f_l};
		\For{\textbf{all} training tracklet $X_i$ of the batch set $X^{it}$}
		\State Calculate $\frac{\partial f_{CE}(W^{it}; X^{it}_i)}{\partial W^{it}_j}$, by back propagation;
		\EndFor
		\State Calculate $\frac{\partial f_L (W^{it}; X^{it})}{\partial W^{it}_j}$ according to Eq. \ref{eq:loss_softmax_derivative};
		\State Update parameters according to Adagrad method; 
		\EndWhile
	\end{algorithmic}
\end{algorithm}
\begin{equation}
\label{eq:f_l}
f_L=\frac{1}{b}\sum^{B}_{i=1} f_{CE}(W^{it}; X^{it}_i)
\end{equation}
\begin{equation}
\label{eq:f_CE}
f_{CE}(W^{it}; X^{it}_i)=-y_{i,0} \cdot log(\sigma(Z_{i,0})) - (1- y_{i,0}) \cdot log(1-\sigma(Z_{i,0}))
\end{equation}
\begin{equation}
\label{eq:loss_softmax_derivative}
\frac{\partial f_L(W^t;\ X^t)}{\partial W_j^t}=\frac{1}{B} \sum_{i=1}^B\left[\frac{\partial f_{CE}(W^t;X_i^t)}{\partial W_j^t}\right]
\end{equation}

%\begin{equation}
%\label{eq:updating}
%W^{t+1}=W^t - \frac{\alpha}{\sqrt{G^t +\epsilon}} \odot \nabla_Wf_L(W^t,X^t) 
%\end{equation} 

\section{Tracklets Generation}
\label{tracklets}

%As it was explained above, the training inputs are tracklets of previously computed features. 
To create a set of training tracklets, a data creation module has been employed\footnote {The data generation tool has been implemented as a set of C++ functions, which are publicly available under http://github.com/magomezs/dataset\_factory/tree/master/data\_factory\_from\_mot}.%, following the procedure shown by Fig. \ref{fig:tracklets_creation}. 
This tool extracts person images from the MOT17 dataset. However, instead of forming tracklets directly from these images, firstly features are computed from them, and subsequently, the features are used to create the training, validation and test set of tracklets. Each feature, $x_n$, is computed from a image captured in the frame denoted by $f(x_n)$, and corresponds to the identity $ID(x_n)$. 

Five different formulations have been designed to combine the features, and consequently, five different types of tracklets sets,  $X_I$, $X_{II}$, $X_{III}$, $X_{IV}$, $X_V$, have been created and used to train the MS-DoAS network, and the resulting models have been evaluated. 

These formulations are defined by the following equations, where $M$ is the set size, that is its number of tracklets. Every set is formed by the random mixture of two subsets, one is formed by positive, $X^+$, and the other one, by negative samples, $X^-$. 

\begin{figure}[H]
	\centering
	\includegraphics[width=0.67\textwidth]{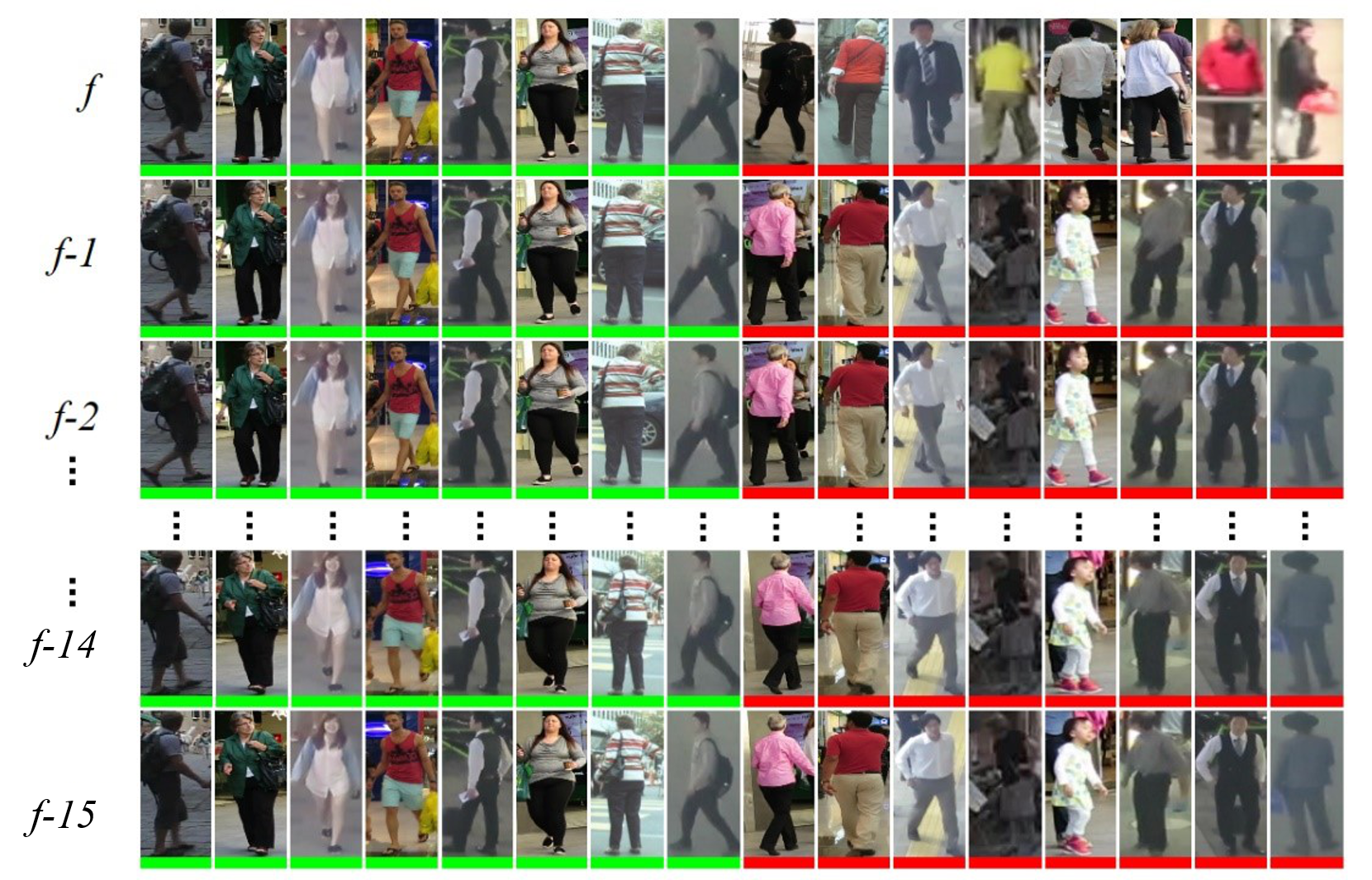}
	\caption{Examples of the images from which tracklets of the set $X_I$ are generated. Positive tracklets are underlined in green, and negatives, in red. $f$ renders the number of the frame from which the first component was extracted, in the sequences of the MOT17 dataset.}
	\label{fig:tracklets}
\end{figure}

The first set, $X_I$, is the simplest one. Every tracklet is formed by features of a certain person in consecutive frames, as Fig. \ref{fig:tracklets} shows. And in the case of the negative tracklets, a different person representation is taken as component $x_0$ to simulate the comparison of an agent with a non-corresponding measurement. The subsets of positive and negative samples for the first set, $X_I^+$ and $X_I^-$, are defined by Eq. \ref{eq:x1+} and \ref{eq:x1-} respectively.

%\begin{figure}[H]
%	\centering
%	\includegraphics[width=0.7\textwidth]{Figs/tracklets_creation.png}
%	\caption{Generation of Feature Tracklets.}
%	\label{fig:tracklets_creation}
%\end{figure}

\begin{multline}
	\label{eq:x1+}
	X_I^+:=Xi=[x_n] \Leftrightarrow \forall n\in Z\cap [0,T-1],f(x_n)-f(x_{n+1})=1 \wedge \\ \forall n\in Z\cap [0,T],ID(x_n)=k \wedge i\leq M   
\end{multline}
\begin{multline}
	\label{eq:x1-}
	X_I^-:=Xi=[x_n] \Leftrightarrow \forall n\in Z\cap [1,T-1],f(x_n)-f(x_{n+1})=1 \wedge \\ \forall n\in Z\cap [1,T],ID(x_n)=k \wedge ID(x_0)\neq k \wedge i\leq M
\end{multline}

The second set, $X_{II}$, is similar to the previous one. However, in positive tracklets, the frame from which component $x_0$ is extracted has not to be consecutive to that for component $x_1$, but a maximum time step (number frames difference) of $F$ frames is allowed between them. In that way, the identification of a person after a short-term occlusion can be simulated to train the model to re-identify agents. The subsets of positive and negative samples for the second set, $X_{II}^+$ and $X_{II}^-$, are defined by Eqs. \ref{eq:x2+} and \ref{eq:x2-}, respectively.

\begin{multline}
	\label{eq:x2+}
	X_{II}^+:=Xi=[x_n] \Leftrightarrow \forall n\in Z\cap [1,T-1],f(x_n)-f(x_{n+1})=1 \wedge \\ f(x_0)-f(x_1)<F \wedge  \forall n\in Z\cap [0,T],ID(x_n)=k \wedge i\leq M
\end{multline}
\begin{equation}
\label{eq:x2-}
X_{II}^-\equiv X_I^-
\end{equation}

The third set, $X_{III}$, allows until $S$ time steps with a maximum size of $F$. Therefore, not only component $x_0$ can be extracted from non-consecutive frames to the adjacent component, but also other randomly located time steps can be generated in both, positive and negative tracklets. In that way, agents\textquotesingle  re-identification in previous frames is simulated. The subsets of positive and negative samples for the third set, $X_{III}^+$ and $X_{III}^-$, are defined by Eqs. \ref{eq:x3+} and \ref{eq:x3-}, respectively. 
\begin{multline}
	\label{eq:x3+}
	X_{III}^+:=Xi=[x_n] \Leftrightarrow \forall n\in Z\cap [0,T-1],f(x_n)-f(x_{n+1})\leq F \wedge \forall n\in \{j\}, \\ \{j\}\subset Z\cap [0,T] \wedge | \{j\} |>(T+1-S) ,f(x_n)-f(x_{n+1})=1\wedge \forall n\in Z\cap [0,T], \\ ID(x_n)=k \wedge i\leq M
\end{multline}
\begin{multline}
	\label{eq:x3-}
	X_{III}^-:=Xi=[x_n] \Leftrightarrow \forall n\in Z\cap [1,T-1],f(x_n)-f(x_{n+1})\leq F \wedge \forall n\in \{j\}, \\ \{j\}\subset Z\cap [0,T] \wedge | \{j\} | >(T+1-S) ,f(x_n)-f(x_{n+1})=1\wedge \forall n\in Z\cap [1,T], \\ ID(x_n)=k \wedge ID(x_0)\neq k \wedge i\leq M
\end{multline}

The fourth set, $X_{IV}$, is similar to the first one but until $N$ intruders are added in random locations of the positive and negative tracklets. That means, that some components are substituted by features of different people, to simulate incorrect associations in previous frames. The subsets of positive and negative samples for the fourth set, $X_{IV}^+$ and $X_{IV}^-$, are defined by \ref{eq:x4+} and \ref{eq:x4-}, respectively, where $\circ$  renders the component-wise product operation. In these equations, $V_i$ renders a binary mask vector, of the same length than the tracklets, randomly generated for every tracklet, $i$. Positions, where the mask takes value $1$, corresponds to the introduction of an intruder. $X'_i$ is a tracklet formed by randomly picked intruders. $V_i$ and $X'_i$ are different for every created tracklet.
\begin{multline}
	\label{eq:x4+}
	X_{IV}^+:=\{ X_i\circ (\neg V_i)+X'_i \circ V_i \mid  V\in Z^{T+1}\wedge V_{i,n}\in Z\cap [0,1] \wedge \sum_{n=0}^T V_{i,n}\leq N \wedge \\ ID(x'_in) \neq ID(x_0)\wedge X_i\in X_I^+ \}
\end{multline}
\begin{multline}
	\label{eq:x4-}
	X_{IV}^-:= \{ X_i\circ (\neg V_i)+X'_i \circ V_i | V_i\in Z^{T+1}\wedge V_{i,n}\in Z\cap [0,1]\wedge \sum_{n=0}^T V_{i,n}\leq N \wedge \\ ID(x'_in)\neq Mo(ID(x_n)) \wedge X_i\in X_I^- \}
\end{multline}

The fifth set, $X_V$, is a combination of the third and the fourth set. It includes time steps and intruders, to train the model with a challenging dataset, making it robust to deal with real inputs when it is applied in a tracking algorithm. The subsets of positive and negative samples for the fifth set, $X_V^+$ and $X_V^-$, are defined by Eqs.\ref{eq:x5+} and \ref{eq:x5-}, respectively. 
\begin{multline}
	\label{eq:x5+}
	X_V^+:=\{ X_i\circ (\neg V_i)+X'_i\circ V_i | V_i\in Z^{T+1}\wedge V_{i,n}\in Z\cap [0,1]\wedge \sum_{n=0}^T V_{i,n}\leq N \wedge \\ ID(x'_in)\neq ID(x_0)\wedge X_i\in X_{III}^+ \}
\end{multline}
\begin{multline}
	\label{eq:x5-}
	X_V^-:= \{X_i\circ (\neg V_i)+X'_i\circ V_i | V_i\in Z^{T+1}\wedge V_{i,n}\in Z\cap [0,1]\wedge \sum_{n=0}^T V_{i,n}\leq N \wedge \\ ID(x'_in)\neq Mo(ID(x_n))\wedge X_i\in X_{III}^- \}
\end{multline}

It should be noted that in the negative tracklets of training sets IV and V, $X_{IV}^-$ and $X_V^-$, the component $x_0$ could present same $id$ as some of the intruders, making the discrimination harder. 

In order to generate a wide variety of tracklets, the intruder components and the $x_0$ component in the negative tracklets has been obtaining not only by taking different person detections from the same sequence but also from different ones, %To avoid the overlapping of the numbers of frames and identities numbers among different sequences, firstly a general list of samples has been generated from all the employed sequences. When a person image of a certain frame is added to the list, the frame number, $f$, and its identification number $ID$ are modified in order to not take the already used values, corresponding to images previously taken from other sequences. In that way, the equations defined above can be applied over this general person images list,
resulting in larger and cross-sequence training sets of tracklets.

\section{Experimental Results}
\label{results}
To evaluate the proposed model, both, its discriminative capacity and the performance of the MOT algorithm where the model has been integrated have been tested. The used dataset and the protocol followed to train and test the model are described below.

\subsection{Datasets}
\label{datasets}

%Multi-Object Tracking datasets are usually composed by a collection of surveillance video-sequences, together with a ground truth file for each one of them. The ground truth file contains the location and identity of every individual at every frame of the sequence. The identity of every person is denoted by an identification number whose value is maintained through the video.

The MOT17\footnote{MOT17 dataset is publicly available under https://motchallenge.net/} dataset has been selected to train and test the model. This dataset belongs to the MOTchallenge\footnote{MOTChallenge is a Multiple Object Tracking Benchmark which provides a unified framework to standardise the evaluation of MOT methods. This is published under https://motchallenge.net/} and was released in $2017$. It contains fourteen variate real-world surveillance video sequences in unconstrained environments (twelve outdoor sequences and two indoor sequences), filmed with both static and moving cameras. It contains the same sequences as MOT16 \cite{milan2016mot16}, but with an extended more accurate ground truth. The resolution is 1920x1080 in twelve of the sequences and 640x480 in the rest of them. There is a total of 11235 frames and 546 different identities. 

The sequences of MOT17 dataset are split into two groups. The sequences of the first group are labelled, i.e. they are accompanied by their ground true files with annotations about individuals' location and identity. Person images have been extracted from this group, and they have been divided, in turn, in two groups to train and test the discrimination capacity of the proposed neural model. 

Secondly, the unlabelled sequences of the second set have been used to evaluate an MOT algorithm where the proposed MS-DoAS model has been used. 
Since the ground truth of these sequences is not publicly available, the algorithm's output has been submitted to the public evaluation platform of the MOT Challenge, which provides the results of calculating standard performance metrics.

% evaluation process is performed by submitting the  , where the comparison with the ground truth is made and  are provided. 

Furthermore, for every sequence, MOT17 dataset provides the detections given by three different people detectors (DPM\cite{felzenszwalb2010object}, Faster-R-CNN \cite{ren2015faster}, and SDP \cite{yang2016exploit}). Therefore, three different versions of every sequence are available, resulting in $42$ sequences.

\subsection{Evaluation of the discriminative capacity of the model}
\label{similarity_evaluation}
This article proposes measuring the Degree of Appearance Similarity through a pair-wise binary classification model. % A model has been trained over each one of the five different generated training sets. %, formed by samples of person images from  {MOT} sequences. 
This has been evaluated as a binary classifier, in order to test its performance to discriminate between positive and negative tracklets, which is rendered by a ROC curve \cite{hanley1982meaning}. 

This curve illustrates the diagnostic ability of a binary classifier as its discrimination threshold, $th$, is varied. 
%A test set of tracklets of images has been generated\footnote{The data generation tool has been implemented as a set of C++ functions, which are publicly available under: http://github.com/magomezs/dataset\_factory/tree/master/data\_factory\_from\_mot.}, with images of multiple individuals from sequences.  
%The performance of the learnt models to properly classify the samples of the test set has been 
%The output of the proposed model is the MS-DoAS. 
$th$ defines the value until which the classifier output, MS-DoAS metric, is considered as the prediction of a negative tracklet, and from which it is considered as a positive tracklet. In that way, the chosen threshold, $th$, divides the distance space in two ranges of values corresponding to each class. 

The ROC curve plots the True Positive Rate ($TPR$) against the False Positive Rate ($FPR$), defined by Eqs. \ref{eq:tpr} and \ref{eq:fpr}, respectively, where $TP$, $TN$, $FP$ and $FN$ are the number of true positives, true negatives, false positives and false negatives, respectively. 

\begin{equation}
\label{eq:tpr}
TPR=\frac{TP}{TP+FN}
\end{equation}

\begin{equation}
\label{eq:fpr}
FPR=\frac{FP}{FP+TN}    
\end{equation}

Moreover, $F1$ score, defined by Eq. \ref{eq:f1} provides a trade-off between the Positive Predictive Value, ($PPV$), Eq. \ref{eq:ppv}, and the True Positive Rate ($TPR$), Eq. \ref{eq:tpr}. For that reason $F1$ has been used to compare methods in the conducted evaluation, as well as, the Accuracy metric, $A$, defined by Eq. \ref{eq:accuracy}, which is the proportion of well-classified samples. 
The number of positive and negatives samples in the test set has been completely balanced to provide a fair evaluation through the accuracy metric, which is not appropriate for the case of having skewed classes. 

\begin{equation}
\label{eq:ppv}
PPV=\frac{TP}{TP+FP}    
\end{equation}

\begin{equation}
\label{eq:f1}
F1=\frac{2 \cdot PPV \cdot TPR}{ PPV+TPR }  
\end{equation}

\begin{equation}
\label{eq:accuracy}
A=\frac{TP+TN}{TP+FP+TN+FN}
\end{equation}

Five different formulations have been designed to generate training tracklets that simulate the tracking of the following type of agents:

\begin{enumerate}[I.]
	\item People who have been previously well identified.
	\item Just re-identified people after a temporary disappearance.
	\item People who have suffered from several disappearances, i.e. their tracks have been interrupted several times.
	\item People who have been wrongly identified (mismatched) in some of the previous frames.
	\item People who have suffered from several disappearances and mismatches.
\end{enumerate}
%-------------------------------------------

%From each formulation, a training set has been created, which is able to train the model to cope with certain situations, according to the characteristics of its tracklets (see Section \ref{tracklets}).

Five different experiments have been conducted. They are called Exp.MS-DoAS.i, where i takes value 1, 2, 3, 4, or 5 to denote that the model has been trained on the set TR1, TR2, TR3, TR4, or TR5, respectively.
Every experiment provides a model that has been tested over five different test sets TS1, TS2, TS3, TS4, TS5, which were also generated according to the five presented tracklets formulations. Therefore $25$ different tests have been performed. %Each experiment results in a model trained over each one of the generated tracklets sets, allowing the evaluation of their effects on the model performance. % Table \ref{tab:exp_tracklets} presents the experiments settings. 

%\begin{table}[H]
%	\centering
%	\caption{Experiment settings to evaluate the effects of five different training sets of tracklets.}
%	\label{tab:exp_tracklets}       % Give a unique label
%	\begin{tabular}{|p{1.9cm}|p{4.4cm}|p{4cm}|}
%		\cline{1-3}
%		\multirow{2}{*}{Experiments} & \multicolumn{2}{l|}{Description}   \\ \cline{2-3}
%		& Training set  & Other settings                    \\ \cline{1-2}
%		\hline
%		\hline
%		Exp.MS\_DoAS.1 &  TR1, from formulation I, Eqs. \ref{eq:x1+} and \ref{eq:x1-} &\multirow{2}{5.5cm}{\parbox[t]{4cm}{LSTM learning \\ VGG11 pre-trained features.  \\ Without data augmentation. \\  L2 regularization, $wd=0.0005$. }} \\ \cline{1-2}
%		Exp.MS\_DoAS.2  & TR2, from formulation II, Eqs. \ref{eq:x2+} and \ref{eq:x2-}& \\ \cline{1-2}	
%		Exp.MS\_DoAS.3	& TR3, from formulation III, Eqs. \ref{eq:x3+} and \ref{eq:x3-}& \\ \cline{1-2}	
%		Exp.MS\_DoAS.4	& TR4, from formulation IV, Eqs. \ref{eq:x4+} and \ref{eq:x4-} & \\ \cline{1-2}	
%		Exp.MS\_DoAS.5	& TR5, from formulation V, Eqs. \ref{eq:x5+} and \ref{eq:x5-}	& \\ \cline{1-3}
%	\end{tabular}
%\end{table}

\begin{figure}[H]
	\centering
	\includegraphics[width=0.9\textwidth]{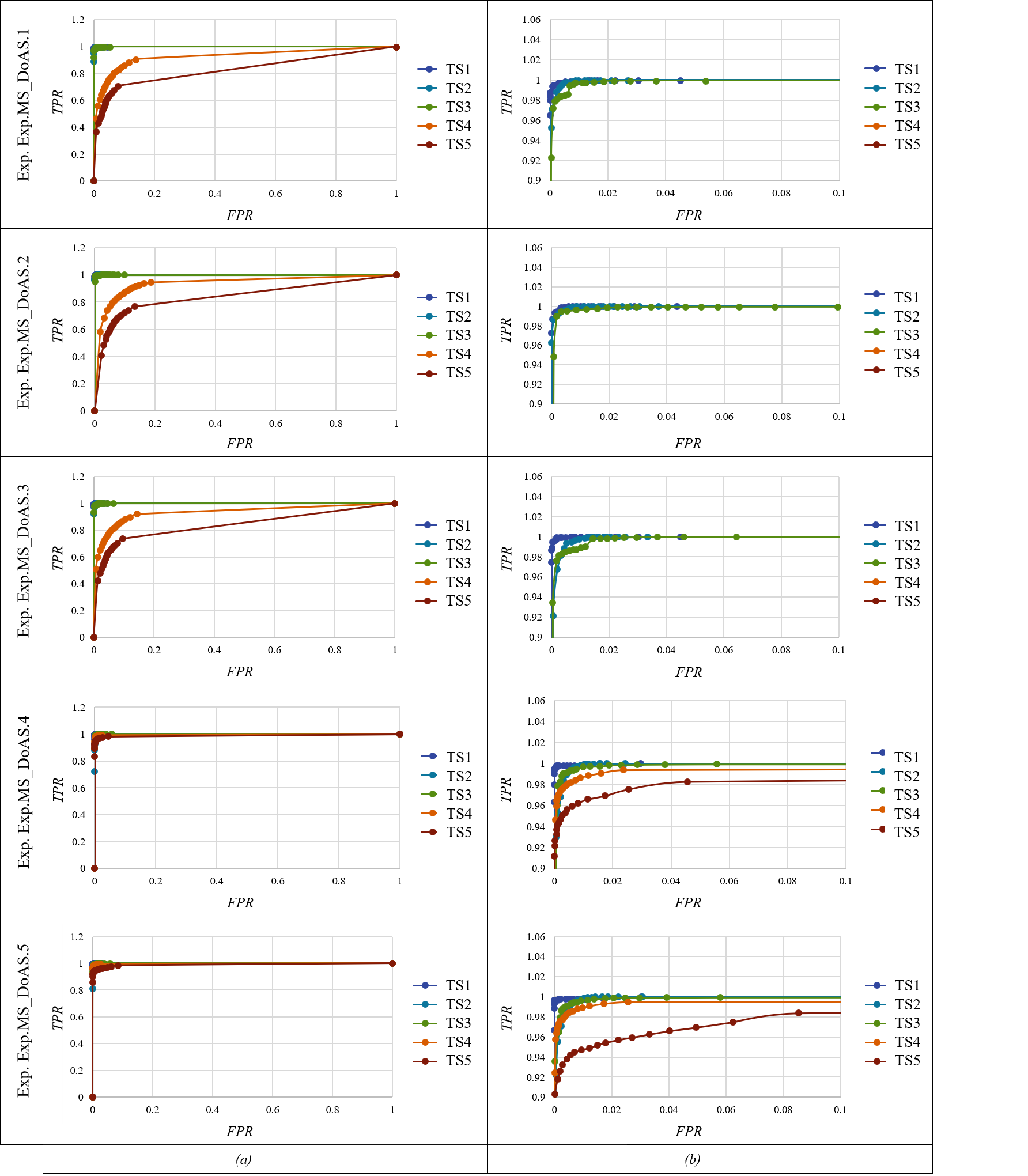}
	\caption{ {ROC} curves from the evaluation of every  {MS-DoAS} model on five different test sets (a), and zoomed region (b).}
	\label{fig:ms_doas_exp}
\end{figure}

\begin{figure}[H]
	\centering
	\includegraphics[width=0.95\textwidth]{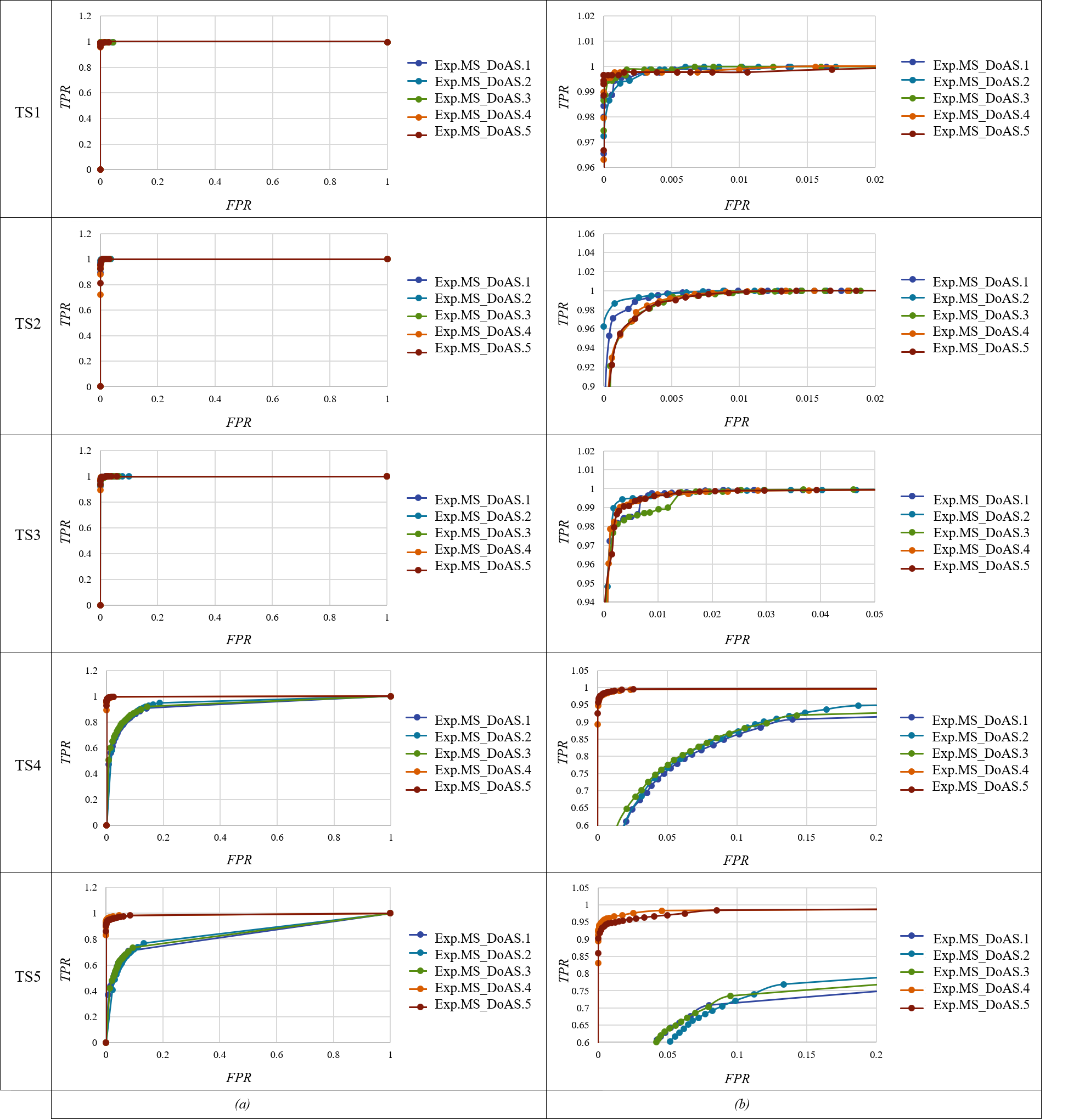}
	\caption{ {ROC} curves from five different evaluations conducted for each one of the  {MS-DoAS} models (a), and zoomed region (b).}
	\label{fig:ms_doas_ts}
\end{figure}

Fig. \ref{fig:ms_doas_exp} presents a comparative graphic for every obtained MS-DoAS model. 
Each graphic shows the ROC curves resulting from testing the query model over the five test sets. 
On the other hand, Fig. \ref{fig:ms_doas_ts} presents a comparative graphic for every test set. Each graphic shows the ROC curves resulting from testing every obtained model over the query test set. Moreover, tables \ref{tab:f1_msdoas_model} and \ref{tab:accuracy_msdoas} show the highest values of the $F1$ score and the accuracy metric, $A$, for each one of the $25$ conducted tests. These tables also provide the classification threshold, $th$, with which the maximum values were achieved. 

\begin{table}[H]
	\centering
	\caption{Maximum $F1$ score value (in [\%]) for every  {MS-DoAS} model, evaluated over five different test sets, and $th$ value where it is achieved.}
	\label{tab:f1_msdoas_model}       % Give a unique label
	\begin{tabular}{l|c|c|c|c|c|}
		\cline{2-6}
		&	TS1&	TS2&	TS3&	TS4&	TS5 \\ \cline{2-6} \hline
		\multicolumn{1}{|c||}{Exp.MS\_DoAS.1} &	$\cca{99.74}$ &	$\cca{99.63}$ &	$\cca{99.43}$ &	$\cca{85.37}$ &	$\cca{77.57}$ \\
		\multicolumn{1}{|c||}{} &	($th$=$0.45$)&	($th$=$0.45$) &	($th$=$0.55$)	& ($th$=$0.9$) &	($th$=$0.95$) \\ \hline
		\multicolumn{1}{|c||}{Exp.MS\_DoAS.2} &	$\cca{99.78}$ &	$\cca{99.6}$ &	$\cca{99.54}$ &	$\cca{86.17}$ &	$\cca{78.24}$ \\
		\multicolumn{1}{|c||}{} &	($th$=$0.3$)	& ($th$=$0.3$) &	($th$=$0.15$)	& ($th$=$0.7$) &	($th$=$0.95$) \\ \hline
		\multicolumn{1}{|c||}{Exp.MS\_DoAS.3}	& $\cca{99.86}$	& $\cca{99.42}$ &	$\cca{99.19}$ &	$\cca{86.01}$ &	$\cca{78.38}$ \\
		\multicolumn{1}{|c||}{} &	($th$=$0.4$)	& ($th$=$0.25$) &	($th$=$0.55$) &	($th$=$0.85$)	& ($th$=$0.95$) \\ \hline
		\multicolumn{1}{|c||}{Exp.MS\_DoAS.4}	& $\cca{99.85}$ &	$\cca{99.49}$ &	$\cca{99.39}$ &	$\cca{98.66}$	& $\cca{97.49}$ \\
		\multicolumn{1}{|c||}{} &	($th$=0.5) &	($th$=$0.65$)	& ($th$=$0.45$)	& ($th$=$0.65$)	& ($th$=$0.75$) \\ \hline
		\multicolumn{1}{|c||}{Exp.MS\_DoAS.5} &	$\cca{99.84}$ &	$\cca{99.43}$	& $\cca{99.39}$ &	$\cca{98.8}$ &	$\cca{96.65}$ \\
		\multicolumn{1}{|c||}{} &	($th$=$0.35$) &	($th$=$0.5$)	& ($th$=$0.45$) &	($th$=$0.65$)	& ($th$=$0.4$) \\ \hline
	\end{tabular}
\end{table}

\begin{table}[H]
	\centering
	\caption{Maximum Accuracy, $A$, value (in [\%]) for every  {MS-DoAS} model, evaluated over five different test sets, and $th$ value where it is achieved.}
	\label{tab:accuracy_msdoas}       % Give a unique label
	\begin{tabular}{l|c|c|c|c|c|}
		\cline{2-6}
		&	TS1&	TS2&	TS3&	TS4&	TS5 \\ \cline{2-6} \hline
		\multicolumn{1}{|c||}{Exp.MS\_DoAS.1} &	 $\cca{99.74}$	& $\cca{99.63}$	& $\cca{99.43}$	& $\cca{88.56}$ &	$\cca{83.43}$\\
		\multicolumn{1}{|c||}{} &	($th$=$0.45$)	& ($th$=$0.45$)	& ($th$=$0.55$) &	($th$=$0.8$) &	($th$=$0.95$)	 \\ \hline
		\multicolumn{1}{|c||}{Exp.MS\_DoAS.2} &	$\cca{99.78}$ &	$\cca{99.60}$	& $\cca{99.54}$ &	$\cca{88.92}$ &	$\cca{82.83}$ \\
		\multicolumn{1}{|c||}{} &	($th$=$0.3$) &	($th$=$0.3$) &	($th$=$0.15$) &	($th$=$0.4$) &	($th$=$0.85$)	 \\ \hline
		\multicolumn{1}{|c||}{Exp.MS\_DoAS.3}	&	$\cca{99.86}$	& $\cca{99.42}$ &	$\cca{99.19}$ &	$\cca{89.09}$ &	$\cca{83.59}$ \\
		\multicolumn{1}{|c||}{} &		($th$=$0.4$) &	($th$=$0.25$)	& ($th$=$0.55$) &	($th$=$0.75$) &	($th$=$0.95$) \\ \hline
		\multicolumn{1}{|c||}{Exp.MS\_DoAS.4}	&$\cca{99.85}$ &	$\cca{99.49}$ &	$\cca{99.39}$ &	$\cca{98.97}$ &	$\cca{98.00}$  \\
		\multicolumn{1}{|c||}{} &	($th$=$0.5$) &	($th$=$0.65$)	& ($th$=$0.45$) &	($th$=$0.65$)	& ($th$=$0.75$) \\ \hline
		\multicolumn{1}{|c||}{Exp.MS\_DoAS.5} &	$\cca{99.84}$ &	$\cca{99.43}$ &	$\cca{99.39}$ &	$\cca{98.08}$ &	$\cca{97.35}$ \\
		\multicolumn{1}{|c||}{} &	($th$=$0.35$)	& ($th$=$0.50$)	& ($th$=$0.45$) &	($th$=$0.65$) &	($th$=$0.4$) \\ \hline
	\end{tabular}
\end{table}

In general, the classification tests prove the high capacity of the proposed model to discriminate between positive and negative tracklets. 

The tests performed over the test sets from TS1 to TS5 are progressively harder since the tracklets on these tests simulate increasingly challenging situations. For that reason, the results are worse from TS1 to TS5. The models obtained from Exp.MS\_DoAS.4 and 5 provide good results for TS5, so they are capable to deal with real tracking situations.

The models learnt utilizing the training sets from TR1 to TR5 are capable of managing progressively more difficult situations, but in detriment of the performance to deal with the easiest identifications. Therefore, a trade-off solution must be chosen, which is given by the models learnt through Exp.MS\_DoAS.4 and Exp.MS\_DoAS.5.

Before developing the proposed Multi-Shot DoAS model, a Single-Shot version was also trained, where the input was not a tracklet, but a pair of images. This model also employed a previously trained VGG11 neural network to compute the features of the input images. Subsequently, the features were compared by the Euclidean distance. This model achieved an accuracy value of $91.7$\% and a $F1$ score of $91.9$\%, which are considerably good values for a challenging task as appearance identification in a MOT sequence. 

Nevertheless, the performance of this metric has been overcome by the novel MS-DoAS model, thanks to the incorporation of temporary consistency with a Multi-Shot design.

%The proposed model compares one person image with a previously captures tracklet to decide whether both correspond to the same person or not.
%The multi-shot design improves the discrimination performance of the similarity appearance model in comparison with the single-shot approach, which simply compares pairs of images.

\subsection{Evaluation of the Multi-Object Tracking performance}
\label{MOT_evaluation}
A tracking-by-detection algorithm has been implemented, where data association is mainly performed by measuring the MS-DoAS between the people detected at every new frame, and the previously tracked identities. The MS-DoAS model has been trained to deal with complex person identification to obtain robust tracking in variate real-world scenarios.
%When the tracking of a person has recently started, and consequently there is not enough information to provide temporary consistency to the association method, the SS-DoAS metric is used instead of the MS-DoAS. 

The MOT performance of the proposed algorithm has been quantitatively measured over the test sequences of MOT17 by the MOTA (Multi-Object Tracking Accuracy). Its computation combines three error sources: false positives, missed targets and identity switches, as Eq. \ref{eq:mota} describes, where $\overline{FN}$, $\overline{FP}$ and $\overline{IDsw}$ sum the number of false negatives, ${FN}_{f}$, false positives, ${FP}_{f}$, and identification switches, ${IDsw}_{f}$, in Eqs. \ref{eq:fn}, \ref{eq:fp} and \ref{eq:idsw} respectively, at every frame, $f$,  of the dataset. All them are divided by the sum of the number of ground truth objects, $g_f$, at every frame.

\begin{equation}
\label{eq:mota}
MOTA=1-(\overline{FN}+\overline{FP}+\overline{IDsw})    
\end{equation}

\begin{equation}
\label{eq:fn}
\overline{FN}=\frac{\sum_{f}{FN}_f}{\sum_{f} g_f}    
\end{equation}

\begin{equation}
\label{eq:fp}
\overline{FP}=\frac{\sum_{f}{FP}_f}{\sum_{f} g_f}    
\end{equation}

\begin{equation}
\label{eq:idsw}
\overline{IDsw}=\frac{\sum_{f}{IDsw}_f}{\sum_{f} g_f}    
\end{equation}

%Furthermore, the  {MOTP}, defined by Eq. \ref{eq:motp}, measures the misalignment between the annotated and the predicted bounding boxes. This is calculated as the sum of intersection areas of each frame, ${IA}_f$, over the union areas ${UA}_f$. ${IA}_f$ is computed as the sum of the intersection area for every identity, this is the common area of its bounding box which is provided by the  {MOT} algorithm and their corresponding ground truth boxes. ${UA}_f$, is the sum of the unions of the areas of those pairs of bounding boxes.

%\begin{equation}
%\label{eq:motp}
%MOTP=\frac{\sum_{f}{IA}_f}{\sum_{f}{UA}_f}
%\end{equation}

Other metrics to measure the  {MOT} performance are:

\begin{itemize}
	\item $ID F1$ Score \cite{ristani2016performance}, which is the ratio of correctly identified detections over the average number of ground-truth and computed detections.
	\item $MT$, Mostly tracked targets. This is the ratio of ground-truth trajectories that are covered by a track hypothesis for at least $80$\% of their respective life span.
	\item $ML$, Mostly lost targets. This is the ratio of ground-truth trajectories that are covered by a track hypothesis for at most $20$\% of their respective life span.
\end{itemize}

Table \ref{tab:mot21} shows the obtained scores for the tracking metrics described above.
This table lists the results for every sequence as well as the global results (marked in bold). The columns corresponding to metrics for which higher scores mean better performance, present white background, and those for which lower scores mean better performance are shaded.

\definecolor{Gray}{gray}{0.9}
\newcolumntype{g}{>{\columncolor{Gray}}c}
\begin{table}[H]
	\centering
	\caption{Multi-Object Tracking performance of algorithm MOT.2 over MOT17 dataset. $MOTA$, $FP$, $FN$, $IDsw$, IDF1, MT and ML values.}
	\begin{tabular}{|l||c||g|g|g||r|r|g}
		\hline
		Sequence        &      $MOTA$       & $FP$              & $FN$               & $IDsw$           &                IDF1 &                 MT & ML                  \\ \hline\hline
		MOT17-01-DPM    &     $23.2$\%      & $414$             & $4,465$            & $74$             &          $ 33.2 $\% &          $ 8.3 $\% & $ 41.7 $\%          \\ \hline
		MOT17-03-DPM    &     $41.7$\%      & $6,280$           & $54,121$           & $595$            &          $ 45.9 $\% &         $ 10.1 $\% & $25.0$\%            \\ \hline
		MOT17-06-DPM    &     $39.1$\%      & $175$             & $6,891$            & $112$            &          $ 43.8 $\% &          $ 5.9 $\% & $ 57.2 $\%          \\ \hline
		MOT17-07-DPM    &     $30.5$\%      & $263$             & $11,356$           & $115$            &          $ 39.1 $\% &          $ 1.7 $\% & $ 48.3 $\%          \\ \hline
		MOT17-08-DPM    &     $16.7$\%      & $592$             & $16,889$           & $116$            &          $ 23.2 $\% &          $ 0.0 $\% & $ 59.2 $\%          \\ \hline
		MOT17-12-DPM    &     $26.7$\%      & $711$             & $5,576$            & $69$             &          $ 42.8 $\% &          $ 2.2 $\% & $ 50.5 $\%          \\ \hline
		MOT17-14-DPM    &     $17.5$\%      & $866$             & $14,238$           & $143$            &          $ 29.8 $\% &            $1.8$\% & $ 62.8 $\%          \\ \hline
		MOT17-01-FRCNN  &     $24.9$\%      & $1,238 $          & $3,563$            & $41$             &          $ 41.0 $\% &         $ 20.8 $\% & $ 33.3 $\%          \\ \hline
		MOT17-03-FRCNN  &     $54.9$\%      & $1,487$           & $45,466$           & $249$            &          $ 55.4 $\% &         $ 22.3 $\% & $  18.9 $\%         \\ \hline
		MOT17-06-FRCNN  &     $47.1$\%      & $330$             & $5,808$            & $99$             &          $ 49.6 $\% &         $ 10.4 $\% & $ 38.7 $\%          \\ \hline
		MOT17-07-FRCNN  &     $29.5$\%      & $770$             & $10,979$           & $153$            &          $ 37.9 $\% &          $ 3.3 $\% & $ 38.3 $\%          \\ \hline
		MOT17-08-FRCNN  &     $18.6$\%      & $576$             & $16,527$           & $97$             &          $ 24.3 $\% &          $ 2.6 $\% & $ 57.9 $\%          \\ \hline
		MOT17-12-FRCNN  &     $34.2$\%      & $332$             & $5,344$            & $23$             &          $ 48.5 $\% &          $ 9.9 $\% & $ 57.1 $\%          \\ \hline
		MOT17-14-FRCNN  &     $18.4$\%      & $1,535$           & $13,321$           & $221$            &          $ 30.7 $\% &          $ 3.0 $\% & $ 56.1 $\%          \\ \hline
		MOT17-01-SDP    &     $36.0$\%      & $692$             & $3,321$            & $118$            &          $ 38.6 $\% &         $ 25. $0\% & $ 29.2 $\%          \\ \hline
		MOT17-03-SDP    &     $64.6$\%      & $2,312$           & $34,207$           & $587$            &          $ 59.7 $\% &         $ 31.8 $\% & $ 14.2 $\%          \\ \hline
		MOT17-06-SDP    &     $49.0$\%      & $361$             & $5,527$            & $117$            &          $ 48.5 $\% &         $ 12.6 $\% & $ 40.5 $\%          \\ \hline
		MOT17-07-SDP    &     $37.7$\%      & $475$             & $9,838$            & $217$            &            $44.5$\% &           $ 6.7$\% & $ 36.7 $\%          \\ \hline
		MOT17-08-SDP    &     $23.2$\%      & $194$             & $15,880$           & $159$            &          $ 27. $1\% &            $5.3$\% & $ 53.9 $\%          \\ \hline
		MOT17-12-SDP    &     $35.3$\%      & $371$             & $5,187$            & $48$             &          $ 45.3 $\% &          $ 8.8 $\% & $ 53.8 $\%          \\ \hline
		MOT17-14-SDP    &     $26.3$\%      & $1,061$           & $12,293$           & $263$            &          $ 36.0 $\% &          $ 1.8 $\% & $ 48.2 $\%          \\ \hline
		\textbf{Global} & $\mathbf{42.3\%}$ & $\mathbf{21,035}$ & $\mathbf{300,797}$ & $\mathbf{3,616}$ & $ \mathbf{46.8\%} $ & $ \mathbf{9.1\%} $ & $ \mathbf{44.1\%} $ \\ \hline
	\end{tabular}
	\label{tab:mot21}
\end{table}

In addition, the global identification precision $P$, takes value $69.9$\% and the global recall, $R$, $35.2$\%.

More than ninety algorithms participate in the MOT17 challenge with $MOTA$ values from $54.7$\% to $-7.3$\%. Therefore, the proposed MOT.2, whose $MOTA$ value is $42.3$\%, can be considered as a relatively effective and robust algorithm. Moreover, it is necessary to highlight the fact that the same algorithm, with the same hyper-parameters setting, has been evaluated over all the MOT17 sequences, proving that the presented method is versatile under multiple and variate tracking situations. 

Besides, the proposed model allows a frame-by-frame data association, resulting in an online tracking algorithm, i.e. the solution is immediately available with each incoming frame and is not changed at any later time, so that, this approach enables real-time surveillance.   
Conversely, the top algorithms from the challenge, LSST17 \cite{feng2019multi} and DGCT, are based on offline (non-causal) methods that rely on future observations.  

This article presents the designing of a novel contrastive metric, the MS-DoAS. However, the performance of the tracking algorithm depends on all the involved stages: detection, agents' state prediction, data association method, and the used contrastive metric. Promising results have been obtained by uniquely addressing this last part, so the complete MOT algorithm is susceptible to potential enhancements.

\section{Conclusions}
\label{conclusions}

This article presents a novel contrastive Deep Convolutional Neural model to perform Multi-Shot recognition by measuring the Degree of Appearance Similarity, MS-DoAS, between successive detections in a Multi-Object Tracking algorithm. 

The design of a Multi-Shot architecture provides temporary consistency to the appearance model, whose recognition capacity is higher than that of its Single-Shot version.
 
Moreover, the proposed model has been trained to deal with different challenging situations, such as with failed associations from previous frames and crossing agents, preventing from further propagation of the identities mismatches. This has been achieved by simulating that type of situations through the training data. 
The formulation for five tracklets generation mechanisms has been designed and implemented in a software tool that has been publicly delivered. 

Eventually, the MS-DoAS model is able to classify quite challenging tracklets with an excellent performance, which makes it suitable for visually identifying a person in a tracking algorithm. Indeed, the tracking of multiple people by a unique visual appearance model without previous knowledge about the scene has been achieved by integrating the proposed model into a plain Tracking-by-Detection algorithm. Its core is a data association mechanism based on taking the MS-DoAS metric as a matching score between two observations.

%The use of a previously trained model (offline learning) avoids the requirement of a large number of frames to learn a reliable model for every agent, during the tracking task. Therefore, the proposed data association presents the same performance from the beginning of the tracking.

Besides, the tracking method does not depend on future observations, i.e. it is an online tracking algorithm, and temporary consistency is provided by modelling the agents’ appearance with a Long Short-Term Memory neuron.

The designed identification model has been tested over datasets presenting a wide variety and number of people, in outdoors and indoors scenarios, and from fixed and moving cameras, from different perspectives, proving its versatility and robustness against multiple scenarios, and its potential application in upcoming more sophisticated systems.

\section{Acknowledgements}
	Research supported by the Spanish Government through the CICYT projects (TRA2016-78886-C3-1-R and RTI2018-096036-B-C21), Universidad Carlos III of Madrid through  (PEAVAUTO-CM-UC3M)  and the Comunidad de Madrid through SEGVAUTO-4.0-CM (P2018/EMT-4362). We gratefully acknowledge the support of NVIDIA Corporation with the donation of the GPUs used for this research.

\bibliographystyle{unsrt}  
\bibliography{references}  

\begin{thebibliography}{10}

\bibitem{mclaughlin2015enhancing}
Niall McLaughlin, Jesus~Martinez Del~Rincon, and Paul Miller.
\newblock Enhancing linear programming with motion modeling for multi-target
  tracking.
\newblock In {\em Applications of Computer Vision (WACV), 2015 IEEE Winter
  Conference on}, pages 71--77. IEEE, 2015.

\bibitem{yang2016temporal}
Min Yang and Yunde Jia.
\newblock Temporal dynamic appearance modeling for online multi-person
  tracking.
\newblock {\em Computer Vision and Image Understanding}, 153:16--28, 2016.

\bibitem{bernardin2008evaluating}
Keni Bernardin and Rainer Stiefelhagen.
\newblock Evaluating multiple object tracking performance: the clear mot
  metrics.
\newblock {\em EURASIP Journal on Image and Video Processing}, 2008(1):1--10,
  2008.

\bibitem{liu2013tracking}
Jingchen Liu, Peter Carr, Robert~T Collins, and Yanxi Liu.
\newblock Tracking sports players with context-conditioned motion models.
\newblock In {\em Proceedings of the IEEE Conference on Computer Vision and
  Pattern Recognition}, pages 1830--1837, 2013.

\bibitem{meijering2009tracking}
Erik Meijering, Oleh Dzyubachyk, Ihor Smal, and Wiggert~A van Cappellen.
\newblock Tracking in cell and developmental biology.
\newblock In {\em Seminars in cell \& developmental biology}, volume~20, pages
  894--902. Elsevier, 2009.

\bibitem{ess2008mobile}
Andreas Ess, Bastian Leibe, Konrad Schindler, and Luc Van~Gool.
\newblock A mobile vision system for robust multi-person tracking.
\newblock In {\em 2008 IEEE Conference on Computer Vision and Pattern
  Recognition}, pages 1--8. IEEE, 2008.

\bibitem{ess2009improved}
Andreas Ess, Konrad Schindler, Bastian Leibe, and Luc Van~Gool.
\newblock Improved multi-person tracking with active occlusion handling.
\newblock In {\em ICRA Workshop on People Detection and Tracking}, volume~2.
  Citeseer, 2009.

\bibitem{kalman1960new}
Rudolph~Emil Kalman.
\newblock A new approach to linear filtering and prediction problems.
\newblock {\em Journal of basic Engineering}, 82(1):35--45, 1960.

\bibitem{gordon1993novel}
Neil~J Gordon, David~J Salmond, and Adrian~FM Smith.
\newblock Novel approach to nonlinear/non-gaussian bayesian state estimation.
\newblock In {\em IEE Proceedings F (Radar and Signal Processing)}, volume 140,
  pages 107--113. IET, 1993.

\bibitem{bae2014robusttracklet}
Seung-Hwan Bae and Kuk-Jin Yoon.
\newblock Robust online multi-object tracking based on tracklet confidence and
  online discriminative appearance learning.
\newblock In {\em Proceedings of the IEEE conference on computer vision and
  pattern recognition}, pages 1218--1225, 2014.

\bibitem{bae2014robust}
Seung-Hwan Bae and Kuk-Jin Yoon.
\newblock Robust online multiobject tracking with data association and track
  management.
\newblock {\em IEEE transactions on image processing}, 23(7):2820--2833, 2014.

\bibitem{fan2010human}
Jialue Fan, Wei Xu, Ying Wu, and Yihong Gong.
\newblock Human tracking using convolutional neural networks.
\newblock {\em IEEE Transactions on Neural Networks}, 21(10):1610--1623, 2010.

\bibitem{le2016long}
Nam Le, Alexander Heili, and Jean-Marc Odobez.
\newblock Long-term time-sensitive costs for crf-based tracking by detection.
\newblock In {\em European Conference on Computer Vision}, pages 43--51.
  Springer, 2016.

\bibitem{tang2016multi}
Siyu Tang, Bjoern Andres, Mykhaylo Andriluka, and Bernt Schiele.
\newblock Multi-person tracking by multicut and deep matching.
\newblock In {\em European Conference on Computer Vision}, pages 100--111.
  Springer, 2016.

\bibitem{chen2015multitarget}
Xiaojing Chen, Le~An, and Bir Bhanu.
\newblock Multitarget tracking in nonoverlapping cameras using a reference set.
\newblock {\em IEEE Sensors Journal}, 15(5):2692--2704, 2015.

\bibitem{zhang2015tracking}
Shu Zhang, Yingying Zhu, and Amit Roy-Chowdhury.
\newblock Tracking multiple interacting targets in a camera network.
\newblock {\em Computer Vision and Image Understanding}, 134:64--73, 2015.

\bibitem{nam2016learning}
Hyeonseob Nam and Bohyung Han.
\newblock Learning multi-domain convolutional neural networks for visual
  tracking.
\newblock In {\em Proceedings of the IEEE Conference on Computer Vision and
  Pattern Recognition}, pages 4293--4302, 2016.

\bibitem{shu2012part}
Guang Shu, Afshin Dehghan, Omar Oreifej, Emily Hand, and Mubarak Shah.
\newblock Part-based multiple-person tracking with partial occlusion handling.
\newblock In {\em 2012 IEEE Conference on Computer Vision and Pattern
  Recognition}, pages 1815--1821. IEEE, 2012.

\bibitem{held2016learning}
David Held, Sebastian Thrun, and Silvio Savarese.
\newblock Learning to track at 100 fps with deep regression networks.
\newblock In {\em European Conference on Computer Vision}, pages 749--765.
  Springer, 2016.

\bibitem{leal2016learning}
Laura Leal-Taix{\'e}, Cristian Canton-Ferrer, and Konrad Schindler.
\newblock Learning by tracking: Siamese cnn for robust target association.
\newblock In {\em Proceedings of the IEEE Conference on Computer Vision and
  Pattern Recognition Workshops}, pages 33--40, 2016.

\bibitem{zhai2018deep}
Mengyao Zhai, Lei Chen, Greg Mori, and Mehrsan~Javan Roshtkhari.
\newblock Deep learning of appearance models for online object tracking.
\newblock In {\em European Conference on Computer Vision}, pages 681--686.
  Springer, 2018.

\bibitem{kim2015multiple}
Chanho Kim, Fuxin Li, Arridhana Ciptadi, and James~M Rehg.
\newblock Multiple hypothesis tracking revisited.
\newblock In {\em Proceedings of the IEEE International Conference on Computer
  Vision}, pages 4696--4704, 2015.

\bibitem{reid1979algorithm}
Donald Reid.
\newblock An algorithm for tracking multiple targets.
\newblock {\em IEEE transactions on Automatic Control}, 24(6):843--854, 1979.

\bibitem{schulter2017deep}
Samuel Schulter, Paul Vernaza, Wongun Choi, and Manmohan Chandraker.
\newblock Deep network flow for multi-object tracking.
\newblock In {\em Proceedings of the IEEE Conference on Computer Vision and
  Pattern Recognition}, pages 6951--6960, 2017.

\bibitem{shitrit2014multi}
Horesh~Ben Shitrit, J{\'e}r{\^o}me Berclaz, Fran{\c{c}}ois Fleuret, and Pascal
  Fua.
\newblock Multi-commodity network flow for tracking multiple people.
\newblock {\em IEEE transactions on pattern analysis and machine intelligence},
  36(8):1614--1627, 2014.

\bibitem{yang2012online}
Bo~Yang and Ram Nevatia.
\newblock An online learned crf model for multi-target tracking.
\newblock In {\em 2012 IEEE Conference on Computer Vision and Pattern
  Recognition}, pages 2034--2041. IEEE, 2012.

\bibitem{zhang2008global}
Li~Zhang, Yuan Li, and Ramakant Nevatia.
\newblock Global data association for multi-object tracking using network
  flows.
\newblock In {\em Computer Vision and Pattern Recognition, 2008. CVPR 2008.
  IEEE Conference on}, pages 1--8. IEEE, 2008.

\bibitem{shitrit2011tracking}
Horesh~Ben Shitrit, Jerome Berclaz, Francois Fleuret, and Pascal Fua.
\newblock Tracking multiple people under global appearance constraints.
\newblock In {\em 2011 International Conference on Computer Vision}, pages
  137--144. IEEE, 2011.

\bibitem{sadeghian2017tracking}
Amir Sadeghian, Alexandre Alahi, and Silvio Savarese.
\newblock Tracking the untrackable: Learning to track multiple cues with
  long-term dependencies.
\newblock In {\em Proceedings of the IEEE International Conference on Computer
  Vision}, pages 300--311, 2017.

\bibitem{simonyan2014very}
Karen Simonyan and Andrew Zisserman.
\newblock Very deep convolutional networks for large-scale image recognition.
\newblock 2014.

\bibitem{krizhevsky2012imagenet}
Alex Krizhevsky, Ilya Sutskever, and Geoffrey~E Hinton.
\newblock Imagenet classification with deep convolutional neural networks.
\newblock In {\em Advances in neural information processing systems}, pages
  1097--1105, 2012.

\bibitem{gomez2017deep}
Mar{'\i}a~Jos{'e} G{\'o}mez-Silva, Jos{'e}~Mar{'\i}a Armingol, and Arturo de~la
  Escalera.
\newblock Deep part features learning by a normalised double-margin-based
  contrastive loss function for person re-identification.
\newblock In {\em In Proceedings of the 12th International Joint Conference on
  Computer Vision, Imaging and Computer Graphics Theory and Applications
  (VISIGRAPP 2017) (6: VISAPP)}, pages 277--285, 2017.

\bibitem{gomez2019balancing}
Mar{\'\i}a~Jos{\'e} G{\'o}mez-Silva, Jos{\'e}~Mar{\'\i}a Armingol, and Arturo
  de~la Escalera.
\newblock Balancing people re-identification data for deep parts similarity
  learning.
\newblock {\em Journal of Imaging Science and Technology}, 2019.

\bibitem{duchi2011adaptive}
John Duchi, Elad Hazan, and Yoram Singer.
\newblock Adaptive subgradient methods for online learning and stochastic
  optimization.
\newblock {\em Journal of Machine Learning Research}, 12(Jul):2121--2159, 2011.

\bibitem{milan2016mot16}
Anton Milan, Laura Leal-Taix{\'e}, Ian Reid, Stefan Roth, and Konrad Schindler.
\newblock Mot16: A benchmark for multi-object tracking.
\newblock {\em arXiv preprint arXiv:1603.00831}, 2016.

\bibitem{felzenszwalb2010object}
Pedro~F Felzenszwalb, Ross~B Girshick, David McAllester, and Deva Ramanan.
\newblock Object detection with discriminatively trained part-based models.
\newblock {\em IEEE transactions on pattern analysis and machine intelligence},
  32(9):1627--1645, 2010.

\bibitem{ren2015faster}
Shaoqing Ren, Kaiming He, Ross Girshick, and Jian Sun.
\newblock Faster r-cnn: Towards real-time object detection with region proposal
  networks.
\newblock In {\em Advances in neural information processing systems}, pages
  91--99, 2015.

\bibitem{yang2016exploit}
Fan Yang, Wongun Choi, and Yuanqing Lin.
\newblock Exploit all the layers: Fast and accurate cnn object detector with
  scale dependent pooling and cascaded rejection classifiers.
\newblock In {\em Proceedings of the IEEE conference on computer vision and
  pattern recognition}, pages 2129--2137, 2016.

\bibitem{hanley1982meaning}
James~A Hanley and Barbara~J McNeil.
\newblock The meaning and use of the area under a receiver operating
  characteristic (roc) curve.
\newblock {\em Radiology}, 143(1):29--36, 1982.

\bibitem{ristani2016performance}
Ergys Ristani, Francesco Solera, Roger Zou, Rita Cucchiara, and Carlo Tomasi.
\newblock Performance measures and a data set for multi-target, multi-camera
  tracking.
\newblock In {\em European Conference on Computer Vision}, pages 17--35.
  Springer, 2016.

\bibitem{feng2019multi}
Weitao Feng, Zhihao Hu, Wei Wu, Junjie Yan, and Wanli Ouyang.
\newblock Multi-object tracking with multiple cues and switcher-aware
  classification.
\newblock {\em arXiv preprint arXiv:1901.06129}, 2019.

\end{thebibliography}

\end{document}